\documentclass[cameraready]{Interspeech}

\usepackage{graphicx}
\usepackage{adjustbox}
\usepackage{subfigure}
\usepackage{subcaption}
\usepackage{multirow}
\usepackage{multicol}
\usepackage{latexsym}
\usepackage{arydshln}
\usepackage{hyperref}
\usepackage{float}
\usepackage{bm}
\usepackage{xcolor}
\usepackage{xspace}

\title{TriageSim: A Conversational Emergency Triage Simulation Framework from Structured Electronic Health Records}

\author[orcid=0000-0002-8803-0575, correspondingauthor]{Dipankar}{Srirag}
\author[orcid=0000-0002-2185-8570]{Quoc Dung}{Nguyen}
\author[orcid=0000-0003-2200-9703]{Aditya}{Joshi}
\author[orcid=0000-0002-2020-0865]{Padmanesan}{Narasimhan}
\author[orcid=0000-0002-1835-3475]{Salil}{Kanhere}

\address{
    University of New South Wales, Sydney, Australia
}

\email{\{d.srirag, quoc\_dung.nguyen, aditya.joshi, padmanesan, salil.kanhere\}@unsw.edu.au}

\keywords{spoken dialogue, dialogue generation, large language models, emergency department triage}

\begin{document}

\maketitle

\begin{abstract}
Research in emergency triage is restricted to structured electronic health records (EHR) due to regulatory constraints on nurse-patient interactions. We introduce TriageSim, a simulation framework for generating persona-conditioned triage conversations from structured records. TriageSim enables multi-turn nurse-patient interactions with explicit control over disfluency and decision behaviour, producing a corpus of $\sim800$ synthetic transcripts and corresponding audio. We use a combination of automated analysis for linguistic, behavioural and acoustic fidelity alongside manual evaluation for medical fidelity using a random subset of 50 conversations. The utility of the generated corpus is examined via conversational triage classification. We observe modest agreement for acuity levels across three modalities: generated synthetic text, ASR transcripts, and direct audio inputs. We provide the code for TriageSim at \url{https://github.com/dipankarsrirag/triage-sim.git}.
\end{abstract}

\section{Introduction}

Emergency department (ED) triage is a time-critical decision-making process conducted through brief spoken interactions~\cite{jones2010four, mason2012time}. During these encounters, nurses assess symptom severity, identify red flags, and assign an acuity level indicative of the urgency of care. Unlike diagnosis, which aims to identify a disease, triage focuses on risk assessment under conditions of incomplete information and conversational constraints. Despite the centrality of spoken interaction in triage, existing clinical datasets~\cite{mimic-iii, mimic-iv-note, Yim2023} largely capture structured outcomes or post-hoc triage notes. Realistic multi-turn dialogue and corresponding audio recordings are rarely available due to privacy and regulatory constraints, limiting the potential of conversational triage under speech and acoustic variations.  Prior efforts focus on only diagnosis and utilise text-based conversations generated using role-play with human actors~\cite{papadopoulos-korfiatis-etal-2022-primock57}, expert-authored dialogues~\cite{ben-abacha-etal-2023-empirical}, or text-only patient simulators conditioned on structured data~\cite{wang-etal-2024-notechat, kyung2025patientsim}. They fail to incorporate audio synthesis, acoustic variability, or explicit triage decision frameworks relevant to the ED setting

In this paper, we introduce TriageSim, \textbf{a simulation framework for generating ED triage conversations} conditioned on structured EHR data, triage protocols, and simulated personas. Structured cases derived from MIMIC-IV-ED~\cite{mimic-iv-ed} and publicly available pedagogical materials are used as latent clinical states from which multi-turn nurse-patient interactions are constructed. The nurse and patient personas are both separately conditioned on relevant attributes (such as country of origin, disfluency rate and so on). We evaluate conversations created using TriageSim along three complementary dimensions: (i) linguistic and behavioural fidelity through quantitative analysis, (ii) acoustic fidelity via intelligibility, speaker consistency and naturalness, and (iii) medical fidelity through expert evaluation of symptom coherence and red-flag identification. We then demonstrate the downstream utility of the generated conversations by performing conversational triage classification on clean text, ASR transcripts, and raw audio. Our findings demonstrate that \textbf{TriageSim produces clinically coherent, interactionally stable conversations with linguistic and behavioural variation}. However, conversational triage classification remains intrinsically challenging. Performance differences between synthetic transcripts, ASR outputs, and direct audio are small, suggesting that reasoning difficulty, rather than transcription noise alone, is the dominant bottleneck.

\begin{figure*}[t!]
    \begin{adjustbox}{width=\linewidth, center}
        \includegraphics[width=\linewidth]{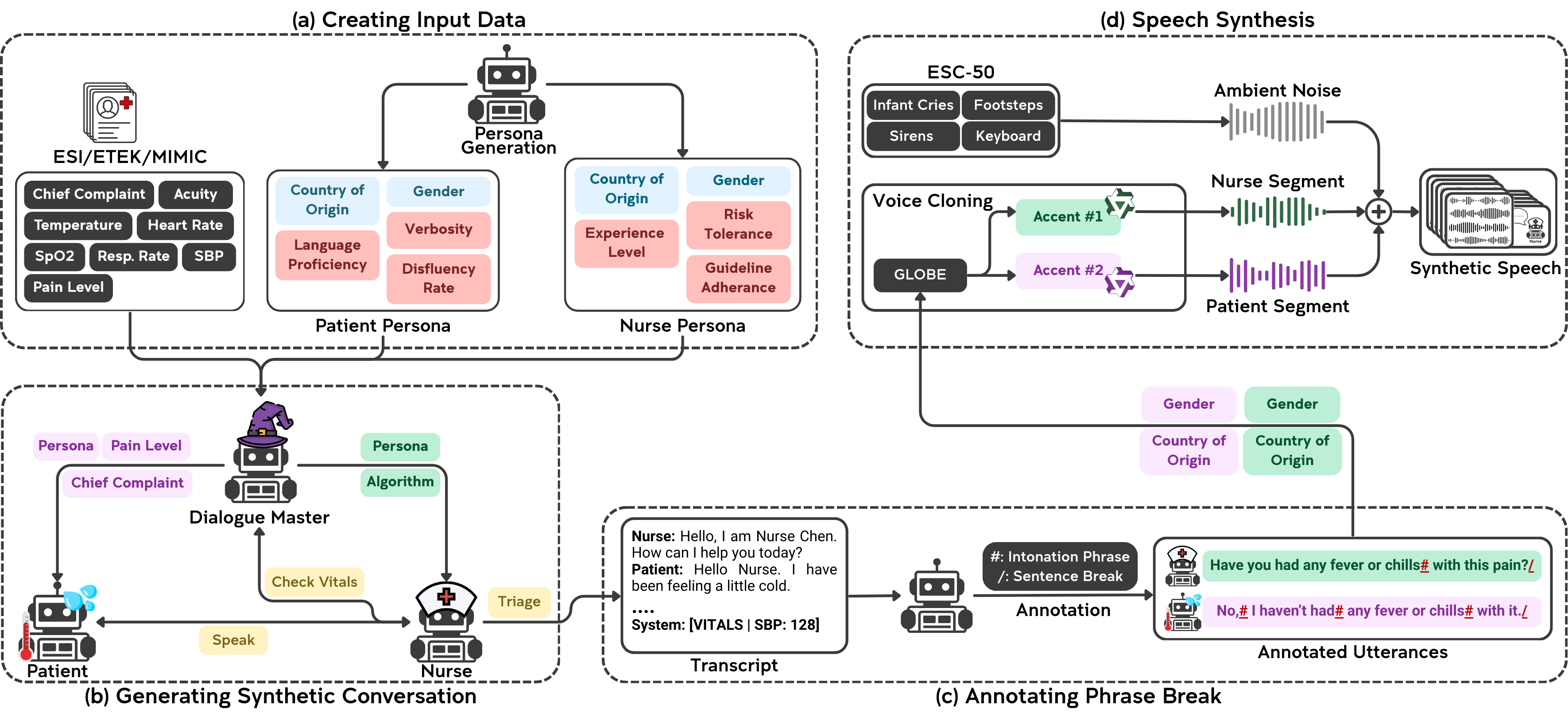}
    \end{adjustbox}
    \caption{Architecture of TriageSim: (a) Input data is constructed using the structured seed data and persona generation. (b) A multi-agent dialogue simulation generates turn-by-turn interaction. (c) Utterances are annotated with phrase structure. (d) Speech is synthesised via accent-conditioned voice cloning and mixed with controlled ambient noise. }
    \label{fig:framework}
\end{figure*}

\section{TriageSim Architecture}
\label{sec:method}

\begin{table}[t!]
    \begin{adjustbox}{width=\linewidth, center}
    \begin{tabular}{cccccc}
        \toprule
        Acuity & Di. & Sp. &  Ut.  & Tokens & Hours \\
        \midrule[\heavyrulewidth]
        1 & 73  & 69 & 9.95 & 21K  & 2.02 \\
        2 & 231 & 138 & 11.46 & 78K  & 7.30 \\
        3 & 309 & 147 & 12.21 & 109K & 10.37 \\
        4 & 112 & 108 & 11.95 & 42K  & 3.68 \\
        5 & 89  & 83 & 12.07 & 30K  & 2.77 \\\hdashline
        Total & 814 & 150 & 11.52 & 280K & 26.14\\
        \bottomrule
    \end{tabular}
    \end{adjustbox}
    \caption{Corpus statistics for each acuity level. \textit{Di.}, \textit{Sp.} and \textit{Ut.} represent \#conversations, \#speakers, \#utterances respectively. Level 1 cases are expected to be shorter than other acuity levels.}
    \label{tab:data-stat}
\end{table}

Figure~\ref{fig:framework} describes the overall architecture of the proposed framework, described as follows. To \textbf{create input data}, we start with primary seed data from ED records~\cite{mimic-iv-ed} containing de-identified triage-level variables including chief complaint, vital signs, pain score, and acuity. To diversify clinical scenarios and mitigate licensing constraints, we incorporate triage cases from Emergency Severity Index (ESI) Handbook~\cite{esi} and the Emergency Triage Education Kit (ETEK) Manual~\cite{etek}, which are standard pedagogical resources from USA and Australia. Free-text scenarios are converted by an ED clinician into structured representations aligned with the MIMIC schema. Following that, we generate structured personas for both the patient and nurse agents, using \texttt{Gemini-3-Pro} (Gemini) and \texttt{GPT-5.2-Pro} (GPT-5), similar to past work~\cite{zhou2024sotopia}. Patient personas include demographic attributes (gender, country of origin, age group), communication traits (language proficiency, verbosity, disfluency rate), and cognitive traits (recall accuracy, cognitive state, topic drift). These attributes influence lexical choice, hesitation patterns, response length, and discourse coherence. Nurse personas include gender, experience level, risk tolerance, and guideline adherence. These attributes modulate question framing, vital sign querying behaviour, triage aggressiveness, and decision thresholds. All persona attributes are generated via structured prompting and validated against case metadata to ensure internal consistency (e.g., gender alignment with the vignette and language proficiency coherence with the disfluency rate).

\begin{figure}[t!]
    \begin{adjustbox}{width=\linewidth, center}      
    \includegraphics{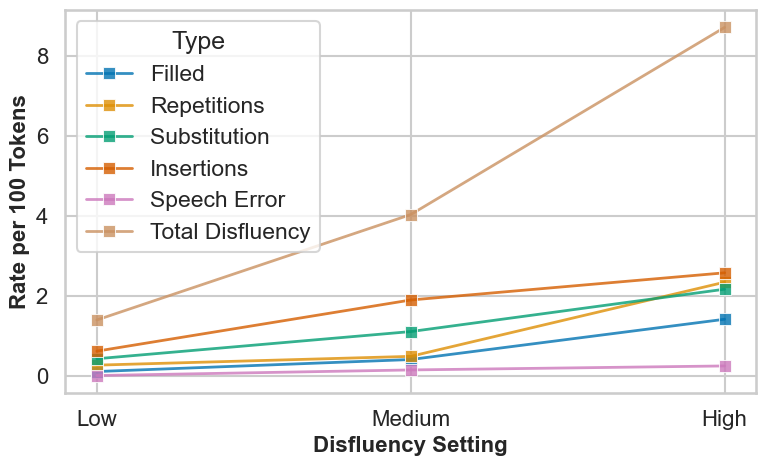}
    \end{adjustbox}
    \caption{
    {Evaluation of Patient disfluency controllability.}
    }
    \label{fig:disfluency_control}
\end{figure}

\begin{figure*}[t!]
    \begin{adjustbox}{width=\linewidth, center}

    \includegraphics[width=0.3\linewidth]{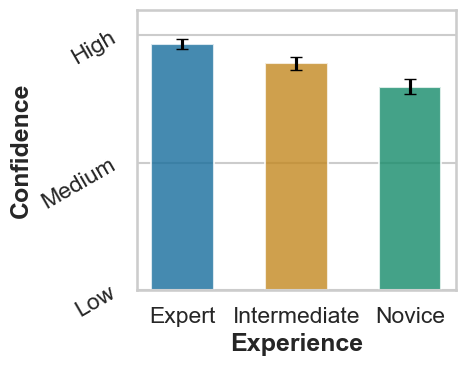}
    \hfill
    \includegraphics[width=0.3\linewidth]{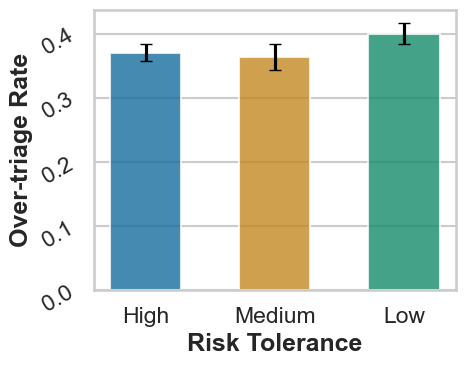}
    \hfill
    \includegraphics[width=0.3\linewidth]{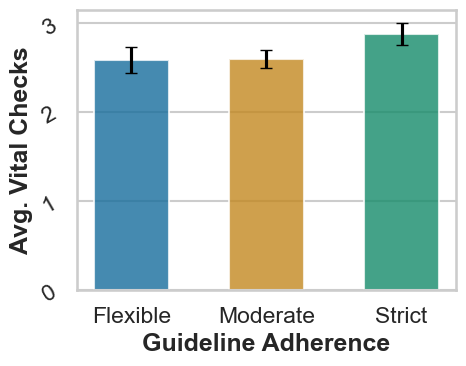}

    \end{adjustbox}

    \caption{
Evaluation of nurse behavioural controllability.
\textbf{Left:} Effect of experience level on triage confidence.
\textbf{Center:} Effect of risk tolerance on over-triage rate.
\textbf{Right:} Effect of guideline adherence on vital-sign checking frequency.
Error bars denote 95\% confidence intervals.
}

    \label{fig:nurse_behaviour}
\end{figure*}

To \textbf{generate synthetic conversations}, three agents work in tandem: a dialogue master, a patient agent, and a nurse agent, each of which are LLMs. The dialogue master has access to the full structured ground truth and enforces case consistency. It instantiates the nurse and patient agents, returns vital measurements when requested and records intermediate triage decisions\footnote{Dialogue master does not expose ground-truth variables to Nurse agent unless explicitly requested via a vital-sign query.}. The nurse agent is provided with a persona specification and a triage algorithm. The nurse agent can either: (a) speak to the patient; (b) request a vital measurement from the dialogue master; (c) perform triage; (d) log any identified medical red flags, indicative acuity level, and confidence at every turn. The nurse operates under either the Australasian Triage Scale~\cite{ESPEJO202557} (ATS) or Emergency Severity Index~\cite{chamberlain2015identification} (ESI). These algorithms, provided as a structured prompt, are implemented as decision policies over acuity criteria, high-risk indicators (red flags) and so on. The nurse agent, therefore, performs triage reasoning constrained by explicit algorithmic guidelines rather than unconstrained free-form generation. In contrast, the patient agent is provided with its persona, chief complaint, and pain level, but lacks access to structured vitals or the indicative symptoms associated with a chief complaint. The interaction proceeds turn-by-turn until a triage decision is made or a maximum turn limit is reached. 

\noindent\textbf{Annotating Phrase Breaks \& Speech Synthesis}: The individual utterances are then annotated with phrase boundaries to provide explicit prosodic control during synthesis. We utilise the framework of~\cite{lee25c_interspeech} for intonation phrase and sentence break annotation. For patient utterances, persona attributes are incorporated into the annotation to more accurately reflect the speaking style. Subsequently, we select a voice sample from publicly available voice banks~\cite{wang24b_interspeech} according to the country of origin and gender specified in the speaker persona. This voice sample, the corresponding transcript and an instruction about the personality traits are provided to Qwen-3-TTS~\cite{Qwen3-TTS} to perform zero-shot voice cloning. See the released code for more details. To approximate ED acoustic conditions, we add background sound events (keyboard typing, telephone ringing, infant crying, ambulance sirens) sourced from ESC-50~\cite{piczak2015dataset}. Noise tracks are mixed at fixed relative gain levels (ambient bed: -32 dB; event sounds: between -14 dB and -26 dB), simulating ED soundscapes. The resulting corpus provides aligned, synthetic transcripts and audio, enabling controlled evaluation of speech-recognition robustness and conversational triage classification under multimodal conditions.

\section{Quality Evaluation}
Using TriageSim, we generate a corpus of 814 conversations, across four accents (Australian, Chinese, Indian, Middle Eastern), using three state-of-the-art language models: Gemini, GPT-5, and \texttt{Claude-Sonnet-4.5}. For each interaction, the generating model is selected uniformly at random, yielding an approximately balanced distribution over models. We promote diversity in triage reasoning by using ESI for ESI Handbook cases, ATS for ETEK cases, and uniformly sampling between ESI and ATS for MIMIC-IV-ED cases. Table~\ref{tab:data-stat} reports corpus statistics stratified by acuity level. Speaker counts are computed within each acuity subset; the same speaker persona may appear in conversations across multiple acuity levels. The conversation generation cost approximately 1000 USD in API calls. We evaluate the fidelity of the corpus along 3 dimensions:

\noindent\textbf{Linguistic and Behavioural Fidelity}: We evaluate if the personality traits are measurably realised in generated interactions. For patients, we focus on disfluency control, measuring rates per 100 tokens for filled pauses, repetitions, substitutions, insertions, and speech errors~\cite{hassan2025enhancingnaturalnessllmgeneratedutterances}. We detect these disfluencies using rule-based regular expressions over transcripts. Figure~\ref{fig:disfluency_control} shows a monotonic increase in realised disfluency across settings. Spearman analysis confirms a positive association between intended and measured disfluency ($\rho$ = 0.57), with repetitions and filled pauses exhibiting the strongest scaling effects ($\rho$ = 0.52 and 0.33, respectively). Additional evaluation on patient utterances using a pretrained disfluency classifier~\cite{marie-2023-disfluency}, shows similar association between difluency probability and intended control level ($\rho$ = 0.45). Beyond patient-level linguistic variation, we evaluate whether nurse behavioural attributes produce corresponding procedural and decision-level effects. Figure~\ref{fig:nurse_behaviour} summarises the behavioural effects of controllable nurse persona attributes. Experience level shows a monotonic relationship with final triage confidence, with experts exhibiting the highest confidence, suggesting that confidence aligns with the intended experience hierarchy. Risk tolerance produces a minimal shift in triage bias: low risk tolerance is associated with a slightly higher over-triage rate (0.40) compared to other levels, indicating small differences in conservativeness. Guideline adherence primarily affects procedural behaviour, with strict agents performing more vital-sign checks on average (2.88) than other agent types. These results indicate that the framework induces measurable changes to various attributes without producing extreme/unstable shifts.

\noindent\textbf{Acoustic Fidelity}: We follow~\cite{wang25x_interspeech} in computing \textit{intelligibility}, \textit{consistency} and \textit{quality}. Overall \textit{intelligibility} (WER=10.8), computed between normalised synthetic transcripts and normalised \texttt{Whisper-Large-V3-Turbo}~\cite{pmlr-v202-radford23a} (Whisper) outputs, indicates that the synthetic speech is clear and recoverable by a SOTA ASR system. Nurse utterances (5.7) are substantially more intelligible than patient utterances (16.0), reflecting the greater disfluency and accent variation introduced in patient speech. Accented speech introduces meaningful challenges, with Middle Eastern (19.4) and Indian (16.7) accents exhibiting higher error rates than Australian (14.2) and Chinese accents (13.2). Manual inspection of the five highest-WER patient utterances per accent group reveals the following patterns. Short lexical items are sometimes substituted with acoustically similar alternatives, such as \textit{fall} transcribed as \textit{four} in a simulated Chinese English example. Filled pauses may be normalised into articles or lexical tokens (\textit{ehh} to \textit{a/air}), and colloquial expressions can be phonetically smoothed (\textit{righto} to \textit{rito}). Number words may be transcribed as digits (\textit{ten} to \textit{10}). These errors predominantly affect short tokens and hesitations rather than longer semantic content, similar to ~\cite{eisenstein2023md3}. Speaker \textit{consistency} is high across all conditions (99.98), indicating highly stable voice identity across turns. Perceptual \textit{quality}, assessed using UTMOSv2~\cite{utmosv2}, yields an average score of 3.42. The corpus is studio-clean by design with controlled perturbation.

\noindent\textbf{Medical Fidelity}: We randomly sample 50 generated conversations and present them to an expert with 5+ years of experience in ED for blinded evaluation. The expert is provided only with the synthesised transcript to (i) draft a textual chief complaint and (ii) assess the red flags identified by the nurse during the interaction, including false positives and missed red flags. Between the ground-truth and clinician-drafted complaints, the mean cosine similarity computed using \texttt{embeddinggemma-300m-medical} from SentenceTransformers~\cite{reimers-gurevych-2019-sentence} is 0.83 (out of 1.0). Similarly, by treating the nurse agent’s logged red flags as predictions and the clinician’s annotations as the reference, red flag detection achieves 0.94 precision and 0.96 recall. These results indicate strong alignment between generated content and clinician interpretation, supporting the medical fidelity of both patient symptom realisation and red-flag identification as generated by the nurse.

\begin{table}[t!]
    \begin{adjustbox}{width=0.9\linewidth, center}
    \begin{tabular}{lcccc}
    \toprule
        \multirow{2}{*}{Model} & \multicolumn{2}{c}{ATS} & \multicolumn{2}{c}{ESI} \\\cmidrule{2-3}\cmidrule{4-5}
        & Syn. & ASR & Syn. & ASR \\
    \midrule[\heavyrulewidth]
    \multicolumn{5}{l}{\textbf{Text}}\\
        Nemo & 0.27 & 0.24 & 0.21 & 0.19\\
        Qwen & 0.34 & 0.32 & 0.27 & 0.27\\
        Grok & 0.39 & 0.38 & 0.30 & 0.33\\\hdashline
        Mean & 0.33 & 0.31 & 0.26 & 0.25\\
        \midrule[\heavyrulewidth]
        \multicolumn{5}{l}{\textbf{Audio}}\\
        Voxtral & \multicolumn{2}{c}{0.28}  & \multicolumn{2}{c}{0.27} \\
    \bottomrule
    \end{tabular}
    \end{adjustbox}
    \caption{Quadratic Weighted Cohen's $\kappa$ for conversational triage classification under ATS and ESI. Text is evaluated on \texttt{Nemotron-3-Nano-30B}~\cite{nvidia2025nemotron3nanoopen}, \texttt{Qwen3-Next-80B}~\cite{yang2025qwen3technicalreport}, and \texttt{Grok-4.1-Fast}, while audio on \texttt{Voxtral-Small-24B}~\cite{liu2025voxtral}. Syn denotes synthetic textual transcripts, and ASR denotes Whisper-generated transcripts from audio files.}
    \label{tab:triage}
\end{table}

\section{Triage Classification}
To assess the practical utility of TriageSim, we evaluate conversational triage classification across three modalities: synthetic transcripts, ASR transcripts from Whisper, and direct audio. We report quadratic weighted Cohen's $\kappa$ to account for the ordinal structure of ATS and ESI acuity levels. Ground-truth acuity labels are derived from the seed data: ESI Handbook cases are labeled with ESI (63), and ETEK cases are labeled with ATS (57). MIMIC-IV-ED cases are restricted to the subset mapped to ESI (370). Table~\ref{tab:triage} shows fair ordinal agreement across conditions. The differences between synthetic and ASR transcripts are minimal, and direct audio classification achieves comparable agreement, suggesting that modality does not alter overall triage accuracy. We do not evaluate the models used for conversation generation to avoid self-preferential bias. These findings suggest that conversational triage difficulty arises primarily from clinical reasoning.

\section{Conclusion}

TriageSim is a simulation framework that generates persona-conditioned, clinically grounded emergency department triage dialogues from structured EHR seed data with aligned text and audio. It combines algorithm-constrained ATS and ESI reasoning with a multi-agent architecture comprising a dialogue master, nurse agent, and patient agent to produce structured, multi-turn synthetic conversations. The resulting conversation transcripts are annotated for phrase breaks and synthesised to speech conditioned on speaker persona, enabling variation in accent, disfluency, and acoustic conditions. Evaluation demonstrates that the corpus is clinically and linguistically coherent, and acoustically stable. Analysis of high-WER utterances highlight challenges introduced by accent variation and colloquial language use. Our findings from conversational triage classification across synthetic text, ASR transcripts, and audio inputs, indicate the primary bottleneck lies in clinical reasoning on conversations rather than transcription noise. TriageSim enables generation of triage conversations from EHR, and can be used to analyse accent-based performance variation.
\section{Acknowledgements}
Dipankar Srirag is supported by NHMRC Ideas Grant RG241647, awarded to Padmanesan Narasimhan and Aditya Joshi in 2025. 

\section{Use of Generative AI Disclosure}
A privacy-preserving AI tool, approved for use at UNSW, was used to assist with drafting and revising portions of the text. All content was manually revised and verified by the authors.

\bibliographystyle{IEEEtran}
\bibliography{mybib}

\end{document}